\documentclass{WileyMSP-template}

\usepackage{cite}
\usepackage{amsmath,amssymb,amsfonts}
\usepackage{algorithm}
\usepackage{algpseudocode}
\usepackage{graphicx}
\usepackage{textcomp}
\usepackage{booktabs}
\usepackage{caption}
\usepackage{subcaption}
\usepackage{float}
\usepackage{pgfplots}
\usepackage{uri}
\usepackage{dblfloatfix}
\usepackage{xcolor}
\usepackage{bm}
\usepackage{hyperref}
\usepackage{xr}

\newcommand{\bcl}[1]{{\color{orange}{}}}
\newcommand{\hh}[1]{\hat{\bm{#1}}}
\newcommand{\monotonic}[2]{\enspace \prec^{#2}_{\mathcal{#1}} \enspace}

\DeclareMathOperator*{\argmax}{arg\,max}

\newcommand{\Z}{Z}

\newcommand{\X}{X}

\begin{document}

\pagestyle{fancy}
\rhead{\includegraphics[width=2.5cm]{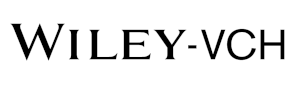}}

\title{PearSAN: A Machine Learning Method for Inverse Design using Pearson Correlated Surrogate Annealing
}

\maketitle


\author{Michael Bezick}
\author{Blake A. Wilson}
\author{Vaishnavi Iyer}
\author{Yuheng Chen}
\author{Vladimir M. Shalaev}
\author{Sabre Kais}
\author{Alexander V. Kildishev}
\author{Brad Lackey*}
\author{Alexandra Boltasseva*}



\begin{affiliations}

Michael Bezick\(^1\), Blake A. Wilson\(^{1}\), Vaishnavi Iyer\(^{1}\), Yuheng Chen\(^{1}\),  
Vladimir M. Shalaev\(^{1}\), Sabre Kais\(^{1}\), Alexander V. Kildishev\(^{1}\), Alexandra Boltasseva\(^{1}\)*, Brad Lackey\(^2\)*

\(^1\) Elmore Family School of Electrical and Computer Engineering, Birck Nanotechnology Center, and  
Purdue Quantum Science and Engineering Institute, Purdue University, West Lafayette, IN, 47907, USA  
\(^2\) Microsoft Quantum, Redmond, WA, USA  

*Corresponding authors: brad.lackey@microsoft.com, aeb@purdue.edu
\end{affiliations}



\begin{abstract}

Inverse design in nanophotonics is often bottlenecked by the curse of dimensionality, particularly for metasurfaces with many degrees of freedom, where the optimization landscape is highly non-convex. In this work, we introduce Pearson Correlated Surrogate Annealing (PearSAN): a machine learning-assisted framework that efficiently navigates these complex spaces. PearSAN uses a surrogate model designed to capture relative performance trends within the latent space of a generative model. This allows the algorithm to rapidly converge towards optimal solutions without requiring exhaustive physical simulations. As a showcase example, PearSAN is applied to the design of thermophotovoltaic (TPV) metasurfaces by matching the working bands of a thermal radiator and a photovoltaic (PV) cell. PearSAN achieves a state-of-the-art design efficiency of $\sim 97\%$, outperforming conventional baselines in both convergence speed and final device performance. These results establish PearSAN as a robust tool for photonics applications requiring precise spectral engineering.

\end{abstract}


\section{Introduction}

With the impressive capabilities of modern generative models, including Generative Adversarial Networks (GANs) \cite{goodfellow_generative_2014, jiang_free-form_2019}, Variational Autoencoders (VAEs) \cite{kingma_auto-encoding_2022}, diffusion models \cite{pabbaraju_provable_2023, ramesh_zero-shot_2021, rombach_high-resolution_2022, Diffussion_2024}, and Large Language Models \cite{vaswani_attention_2017, openai_gpt-4_2023, minaee_large_2024}, there is growing interest in their application across various engineering domains. 
GANs and diffusion models have shown potential for generating device designs with specific properties by integrating surrogate models in place of large and complex simulations.
Surrogate optimization algorithms have become instrumental in accelerating topology optimization and inverse design, where they leverage synthetic training datasets to drastically reduce simulation costs and shorten development cycles \cite{dara_machine_2022, brunner_neural_2023}.

In photonics and optoelectronics, the optimization of new compact planar devices and systems built on metasurfaces is a nascent engineering problem \cite{wilson_authentication_2024, liu_generative_2018, ma_probabilistic_2019, ML-PDD_2025}. Metasurfaces are subwavelength-thin nanostructed films that have fine-tuned control over phase, amplitude, and polarization across various wavelengths through a composition of optimized meta-atoms. Their unique properties have sparked tremendous interest in tailoring metasurfaces for emerging applications in sensing \cite{tabassum_metasurfaces_2022, georgi_metasurface_2019, qin_metasurface_2022}, quantum information \cite{ollanik_integrated_2024}, and renewable energy \cite{elahi_chapter_2023, mascaretti_designing_2023}.

The inverse design of metasurfaces is traditionally limited, however, by the vastness of the geometric parameter space, otherwise known as the curse of dimensionality. As meta-atoms become more intricate to support multi-functional or broadband responses, the number of degrees of freedom scales rapidly. This results in an exponentially large and physically sparse design landscape, making direct optimization impractical \cite{christiansen_inverse_2021}.
Standard topology optimization methods often struggle in these regimes due to the non-convexity of such device design problems, which can trap them in local minima.
One potential direction under exploration to address these limitations is \textit{latent optimization}, a data-driven paradigm that compresses the initial high-dimensional design spaces into lower-dimensional, feature-rich representations known as latent spaces.
Global search algorithms such as simulated annealing \cite{kirkpatrick_optimization_1983}, stochastic gradient descent, Markov Chain Monte Carlo \cite{wilson_non-native_2024} or quantum-inspired optimizers \cite{rajak_quantum_2023} then sample designs through the latent space instead of the higher-dimensional data space. 
This strategy has proven particularly effective in multi-constrained, high-parametric metasurface design: for instance, adversarial autoencoders (AAEs) have utilized this geometric compression to significantly accelerate the discovery of selective thermal emitters \cite{kudyshev_machine-learning-assisted_2020}. Another key advantage of these frameworks is their foundation in variational Bayesian inference, which allows them to approximate the underlying probability distribution of the training data. This structured compression enables the application of sample-efficient frameworks such as Bayesian optimization and active learning to otherwise intractable landscapes in device design problems. By enforcing continuity in the latent space, these models ensure that sampled vectors decode into geometries that faithfully retain the structural motifs and design constraints inherent in the dataset. This concept was further refined using conditional variational autoencoders (cVAEs), which explicitly condition the learned distribution on target performance metrics to optimize nanopatterned integrated photonic components \cite{tang_cvae_2020}.

However, constructing a structured latent manifold is only the first step; efficiently navigating it to find optimal designs requires a guidance mechanism. Thus, some implementations also employ a surrogate model,  a learnable approximation of the Figure-of-Merit (FOM) used to guide optimization within the latent space. To train the surrogate model, early frameworks used binary variational autoencoders with energy matching (EM) as the optimization criterion to generate high-efficiency meta-atom designs that numerically matched the surrogate's learned energy function to the FOM value \cite{wilson_machine_2021}. While these methods made global exploration of the design landscape comparatively more tractable, EM's restrictive nature forces the surrogate function to approximate the FOM, when in reality it only needs to preserve the \textit{ordering} of the designs' FOM. As a result, surrogate model training is over-biased and misleads the optimizer, leading to noisy, weakly correlated, suboptimal designs.

To address these concerns, we introduce Pearson-Correlated Surrogate Annealing (PearSAN). This latent optimization framework employs a Pearson Correlation Surrogate Optimization Loss (PearSOL) to train the surrogate model to preserve ordering rather than accuracy, based on Pearson correlation. PearSOL imposes only a correlated linear relationship between the true FOM and the surrogate's energy function, thereby not restricting the range of values the surrogate model can express, allowing greater expressivity while preserving the relative order of optimality. PearSAN combines PearSOL with variational neural annealing to accelerate sampling of low-energy latent vectors \cite{hibat-allah_variational_2021}. 
PearSOL ensures that the optimizer is more likely to explore regions of the latent space that contain highly performing designs, resulting in higher-quality samples.

We demonstrate PearSAN on the thermophotovoltaic (TPV) metasurface unit-cell design problem by optimizing the operating band of a thermal-emitter metasurface to achieve high overlap with the absorption band of an ideal photovoltaic cell (PV cell). 
The design space of our problem is defined through the material topology of the thermal emitter, which is composed of $TiN$ distributed on a $Si_3N_4$ spacer. In Section 2, we formalize the problem definition and details of the PearSAN framework. We elaborate on the theory underlying isotonic latent optimization, a critical cornerstone of our work. In Section 3, we present our results and provide a rigorous benchmark against established implementations. In Section 4, we conclude with an evaluation of PearSAN and a discussion of future directions for correlation-based inverse design.

\section{Methods}
\subsection{TPV Optimization Setup}
\begin{figure}[t]
    \centering
    \includegraphics[width=\columnwidth]{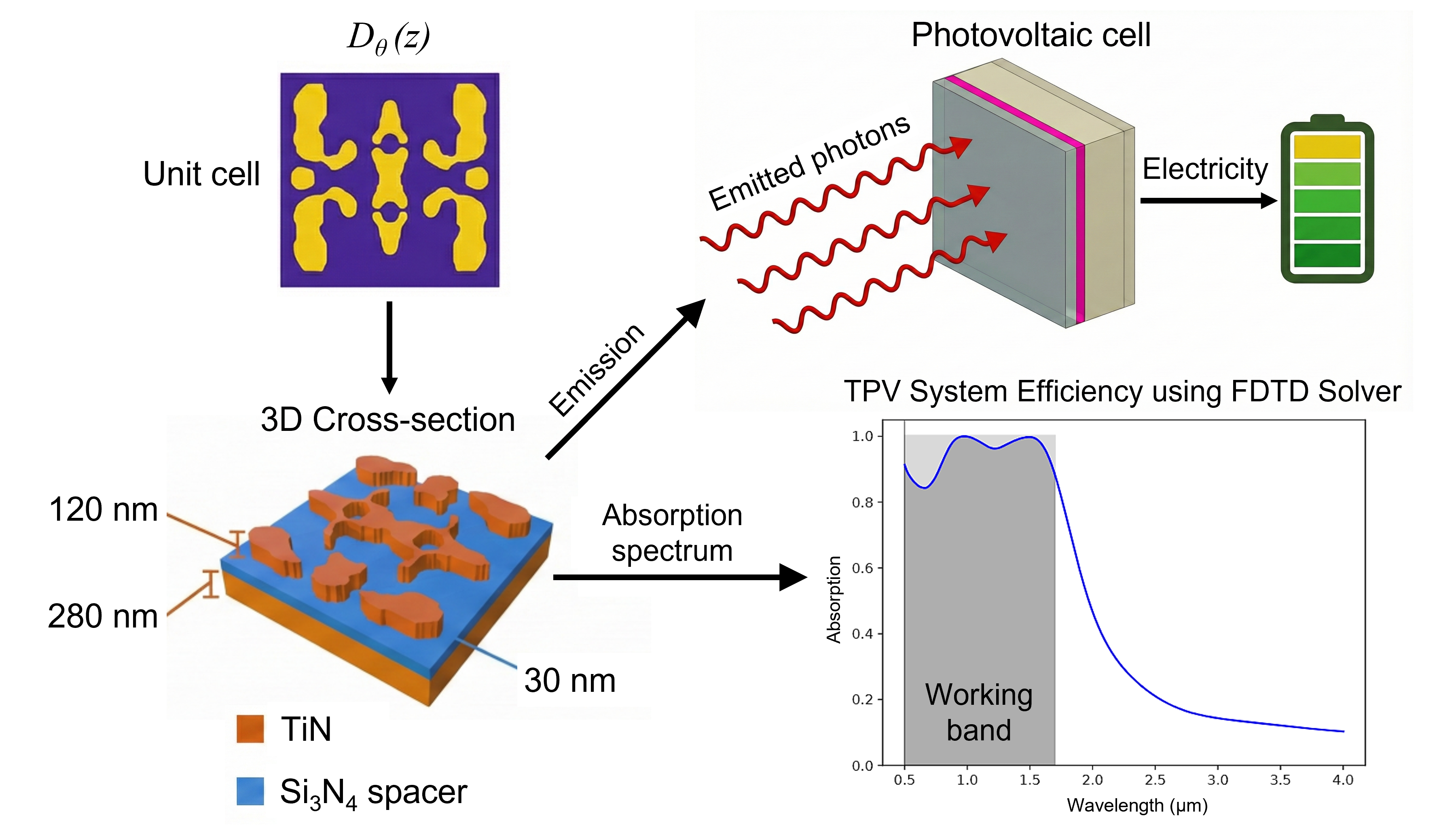} 
    \caption{\textbf{Thermophotovoltaic system: } The TPV system considered in our showcase problem corresponds to the samples $D_{\theta}(\bm{z})$ generated by the pretrained decoder and consists of a thermal emitter and PV cell. The thermal emitter converts thermal energy into photons, and the PV cell generates electricity from these photons within a specific wavelength range (working band: 0.5-$1.7\,\mu\text{m}$, gray area in spectrum subfigure). Each thermal emitter consists of 3 layers: a $120 $ nm titanium nitride plasmonic antenna, a $30 $ nm silicon nitride spacer, and a $280 $ nm titanium nitride back reflector, as shown in the 3D cross-section. Each unit cell is a $280$ nm $\times$ $280$ nm square. The topology of the titanium nitride layer determines the shape of the absorption spectrum, whose near-full absorption range ideally falls solely within the working band of the PV cell. The generated thermal emitter design efficiencies are assessed using a regression-model-based Visual Geometry Group Net (VGGNet) during training and via FDTD simulation after sampling.}
    \label{fig:tpvcells}
\end{figure}

The TPV system (Figure \ref{fig:tpvcells}) relies on a thermal emitter that converts heat into photons, and its performance hinges on engineering the emitter to radiate predominantly within the spectral band where the PV cell can effectively absorb and convert this radiation into electricity \cite{elahi_chapter_2023, sakakibara_practical_2019, khairul_azri_advancement_2023}. In our case, gallium antimonide (GaSb) PV cell has a working band between $\lambda_{\text{min}} = 0.5\, \mu$m and $\lambda_{\text{max}} = 1.7\, \mu$m (indicated as grey area in Figure \ref{fig:tpvcells}). Achieving this requires tailoring the emissivity of the emitter to ensure a significant overlap with the PV cells’ spectral response. The challenge lies in maximizing in-band radiation, which directly contributes to power generation, while minimizing out-of-band radiation that leads to undesirable heating, thereby reducing both the quantum efficiency and the lifespan of the PV cells. The ideal thermal emitter would possess a step-function emissivity profile, characterized by $\varepsilon(\lambda_{\text{min}} \leq \lambda \leq \lambda_{\text{max}}) = 1$ a zero elsewhere. Traditional designs, such as gap-plasmon structures, have been proposed as a means to achieve this spectral reshaping. These structures typically consist of a back reflector, a dielectric spacer, and an array of plasmonic resonators. While such designs are straightforward to fabricate and intuitively simple, they often exhibit limited efficiency due to their basic antenna geometries, which restrict spectral control.

To address these limitations, we explore a more sophisticated emitter design that employs a 280 nm titanium nitride (TiN) back reflector, a 30 nm silicon nitride ($\mathrm{Si}_{3}\mathrm{N}_{4}$) spacer, and a 120 nm top layer composed of TiN plasmonic antennas, with a fixed 280 nm period of the unit cell in both lateral directions (Figure \ref{fig:tpvcells}). The top layer is defined as the optimization region and is discretized with 64 × 64 optimization elements. The primary goal is to optimize the shape and topology of these antennas, to drive the emissivity toward an ideal step-function profile. The efficiency of these optimized thermal emitters is quantified as the product of their in-band and out-of-band efficiencies.

\begin{equation}
\label{eq:eff}
    \text{eff} = \text{eff}^{\text{in}} \cdot \text{eff}^{\text{out}},
\end{equation}

here,

\begin{equation}
\label{eq:effin}
\text{eff}^{\text{in}} = \frac{\int_{\lambda_{\text{min}}}^{\lambda_{\text{max}}} \varepsilon(\lambda) B_\lambda (\lambda, T) \,d\lambda}{\int_{\lambda_{\text{min}}}^{\lambda_{\text{max}}} B_\lambda (\lambda, T) \,d\lambda},
\end{equation}

\begin{equation}
\label{eq:effout}
\text{eff}^{\text{out}} = \frac{\int_{\lambda_{\text{min}}}^{\lambda_{\text{max}}} \varepsilon_{\text{TiN}} (\lambda) B_\lambda (\lambda, T) \,d\lambda}{\int_{\lambda_{\text{min}}}^{\lambda_{\text{max}}} B_\lambda (\lambda, T) \,d\lambda}.
\end{equation}

where the Planck law,

\begin{equation}
\label{eq:B}
B_\lambda (\lambda, T) = \frac{2hc^2}{\lambda^5} \left(\exp\left(\frac{hc}{\lambda k_\text{B} T}\right) - 1\right)^{-1},
\end{equation}

gives the spectral radiance of the blackbody at a given temperature $T$ and wavelength $\lambda$; the fundamental constants include the Planck constant $h$, the Boltzmann constant $k_\text{B}$, and the speed of light in free space $c$. In Eq. \ref{eq:effin} and Eq. \ref{eq:effout}, $\varepsilon(\lambda)$ and $\varepsilon_{\text{TiN}} (\lambda)$ denote the spectral emissivities of the optimized emitter and a bare TiN back reflector, respectively. $T$ is the working temperature of the emitter, and wavelengths $\lambda_{\text{min}}, \lambda_{\text{max}}$ are, respectively, the lower and upper bounds of the PV cell operation band. The in-band efficiency is calculated by normalizing the emitter's radiance within the PV cell's operational band to that of an ideal blackbody emitter at 1800°C. Conversely, the out-of-band efficiency is defined by comparing the radiance outside the operational band against a bare TiN back reflector, recognizing that the long-wavelength response of the gap-plasmon structures is primarily determined by the material properties of TiN, whose optical losses impose fundamental limits on out-of-band emissivity. 

The FOM used in the optimization is defined as
a weighted spectral average of the in-band absorption and the 
out-of-band low absorption. The wavelength–dependent weights are 
taken directly from the ideal emitter’s absorption spectrum 
$A_{\mathrm{id}}(\lambda)$, which represents the desired 
step-function profile. It is defined as a unit-absorption step function within the target working band and zero absorption outside it, thereby specifying the desired spectral profile against which each design is evaluated.

The FOM for a design $x(i)$ is

\begin{equation}
\label{eq:fx}
f(\bm{x}^{(i)}) =
\frac{
\displaystyle
\int A_{\mathrm{id}}(\lambda)\,A(\lambda,\bm{x}^{(i)})\,d\lambda
+
\int \big(1-A_{\mathrm{id}}(\lambda)\big)\,
\big(1-A(\lambda,\bm{x}^{(i)})\big)\,d\lambda
}{
\displaystyle
\int A_{\mathrm{id}}(\lambda)
+
\big(1-A_{\mathrm{id}}(\lambda)\big)\,d\lambda
},
\end{equation}

Here, $A_{\mathrm{id}}(\lambda)$ weights the in-band region so that
$A(\lambda,\bm{x}^{(i)})$ is maximized where the ideal emitter absorbs,
while $1-A_{\mathrm{id}}(\lambda)$ weights the out-of-band region so
that $A(\lambda,\bm{x}^{(i)})$ is minimized, where the ideal emitter should 
not absorb. In this way, the FOM promotes a step-function-like 
absorption spectrum.

\subsection{TPV Dataset Preparation}

For a single TPV unit cell, material distribution within the simulation domain can be viewed as a subset of the overall design parameter space. By leveraging a gradient-based topology optimization (TO) approach in this space, we can converge on optimal, non-intuitive binary layouts that significantly enhance device performance. In this work, we apply TO to TPV emitters to generate a diverse, high-quality training dataset for PearSAN method. Each TPV unit cell is composed of a $64 \times 64$ discretized binary grid, where a value of 1 denotes a raised TiN antenna and 0 corresponds to an air gap, representing the top layer of the thermal emitter design. The initial material distribution is randomly initialized within a single quadrant of the unit cell and then mirrored across the $x$- and $y$-axes to enforce symmetry. This allows us to simulate only one-quarter of the cell while applying symmetry boundary conditions, thus reducing computational cost.

The TO process is implemented using an adjoint optimization method, which requires two full-wave simulations per iteration—one forward and one adjoint—executed via a commercial finite-difference time-domain (FDTD) solver (\textsc{Ansys Lumerical FDTD}). The spatial electromagnetic field distributions from these simulations guide updates to the dielectric structure to maximize a custom-defined FOM. Compared to conventional gap-plasmon structures—typically cylindrical TiN resonators that reach up to 84\% efficiency due to limited spectral tunability—our best TO designs reach up to 92\% efficiency by exploiting complex freeform geometries.

To ensure fabrication feasibility, we apply a three-step robustness algorithm throughout the entire optimization cycle, including the TO process and the post-processing of PearSAN model-generated designs. First, we average the local permittivity to drive intermediate values toward binary states. Second, we average the FOM across perturbed geometries to improve tolerance to fabrication imperfections. Third, spatial filtering is also applied periodically to eliminate sub-30 nm features, and the optimization cycle is limited to 50 iterations. This process consistently produces robust binary structures composed of air and TiN, with superior in-band emission and suppressed out-of-band radiation.

A dataset of 12,000 such TO-generated binary structures is used to train a binary autoencoder (bAE), detailed in Supplementary Information Section S4. This model compresses the binary designs into a low-dimensional latent space while preserving key performance characteristics. The initial latent vector dataset $Z^{(0)}$ is generated by encoding the designs with the bAE and sampling via $\bm{z}^{(i)} \sim \text{Bernoulli}(\mathcal{E}(\bm{x}^{(i)}))$. To efficiently evaluate new candidate designs during generative sampling, we employ a pretrained Visual Geometry Group Net (VGGNet) regression model to approximate the FOM $f(\bm{x}^{(i)})$. We note that the model output is unnormalized and used directly for performance ranking, without scaling the plots to unity.

To extend the use of this TO-generated design apparatus beyond supervised prediction, we interpret TPV emitter design as a latent-space optimization problem. The bAE provides a discrete latent representation in which each binary unit cell $\bm{x}^{(i)}$ is mapped to a code $\bm{z}^{(i)}$ while approximately preserving its FOM $f(\bm{x}^{(i)})$, and the pretrained VGGNet surrogate enables rapid evaluation of candidate designs decoded from these codes. Rather than directly optimizing over the exponentially large space of binary patterns, we therefore focus on learning a generative latent sampler $q_\phi(\bm{z})$ that preferentially explores regions of $\mathcal{Z}$ associated with high-performance emitters under a fixed, optimization-agnostic decoder $p_\theta(\hh{x} \mid \bm{z})$. In the following section, we formalize this strategy as an isotonic latent optimization problem, where samples drawn from $p_\theta$ are encouraged to be probability-ranked in a manner that is monotone with respect to their FOM, and we introduce PearSAN as a practical realization of this framework.

\subsection{Isotonic Latent Optimization }
\label{sec:isola}
Consider a data space $\mathcal{X}$ sampled i.i.d. $\X = \{\bm{x}^{(i)}\}_{i=1}^N$ where each point $\bm{x}^{(i)}$ has a FOM value $f(\bm{x}^{(i)})$. 
Our goal is to train a generative model $p_\theta$ with variational parameters $\theta$ to sample new data $\hh{x} \sim p_\theta(\hh{x})$ that optimizes the expected FOM $f(\cdot)$, i.e.,
\begin{equation}
    \argmax_\theta \mathbb{E}_{\hh{x} \sim p_\theta(\hh{x})}[f(\hh{x})]. \label{eq:opt}
\end{equation}
Since the advent of variational autoencoders \cite{bank_autoencoders_2021, kingma_auto-encoding_2022}, many generative models rely on extracting data features into a low-dimensional {\it latent} space $\mathcal{Z}$ with latent variables $\bm{z} \in \mathcal{Z}$ \cite{rombach_high-resolution_2022}.
Then, the new data is sampled directly from the latent space $\mathcal{Z}$ using a conditional decoder $\hh{x} \sim p_\theta(\hh{x} \vert \bm{z})$.
To leverage latent sampling for optimization, we express $p_\theta(\hh{x}) = \sum_{\bm{z}} p_\theta(\hh{x} \vert \bm{z})q_\phi(\bm{z}) $ with a latent sampler $q_\phi(\bm{z})$ and latent variational parameters $\phi$. Then, we rewrite Eq. \ref{eq:opt} as a latent optimization problem
\begin{equation}
    \argmax_{\theta, \phi} \mathbb{E}_{\bm{z} \sim q_\phi(\bm{z})}[\mathbb{E}_{\hh{x} \sim p_\theta(\hh{x} \vert \bm{z})}[f(\hh{x})]]. \label{eq:latentopt}
\end{equation}
This splits our model $p_\theta$ into a latent optimization sampler $q_\phi$ and a decoder $p_\theta(\hh{x} \vert \bm{z})$. 
Rather than training the decoder to be aware of the optimization distribution of the latent sampler $q_\phi$ \cite{kudyshev_machine_2021}, which can lead to sampling an unnormalized distribution at each gradient step \cite{khoshaman_quantum_2018} and meticulous fine-tuning problems \cite{minaee_large_2024}, we train the latent sampler to produce optimal latent vectors under a pretrained ``optimization-agnostic'' decoder $p_\theta(\hh{x} \vert \bm{z})$. \footnote{In principle, one could apply PearSAN or other latent optimization techniques to any pretrained decoder if its latent space is discrete. Future work will focus on the continuous case. }

The optimal latent sampler $q_\phi$ would solve Eq. \ref{eq:opt} by producing the latent vector for a decoded design with the maximal FOM with probability 1, however this is generally infeasible to accomplish in practice because of the typically exponentially large latent space \cite{wang_binary_2023}. 
Instead, we make only the weak assumption that our model $p_\theta$ produces samples whose probability is (strictly) isotonic\footnote{That is, the larger the FOM the more probable the design.} with the FOM values over the design space.
We say that $p_\theta$ is (strictly) isotonic with $f$ over $\mathcal{X}$, i.e., $f(\hh{x}) \monotonic{X}{+} p_\theta(\hh{x})$, to mean that for each pair of points $(\hh{x}^{(i)}, \hh{x}^{(j)}) \in \mathcal{X} \times \mathcal{X}$ we have 
\footnote{For minimization problems, we use $f(\hh{x}) \monotonic{X}{-} p_\theta(\hh{x})$ to denote antitonic relationships $f(\hh{x}^{(i)}) < f(\hh{x}^{(j)}) \implies p_\theta(\hh{x}^{(i)}) > p_\theta(\hh{x}^{(j)}).$ 
}
\begin{align}
    f(\hh{x}^{(i)}) < f(\hh{x}^{(j)}) \implies p_\theta(\hh{x}^{(i)}) < p_\theta(\hh{x}^{(j)}). \label{eq:mono}
\end{align}
\bcl{Do we want to also introduce the notion of ``strictly'' isotone (i.e. replace $\leq$ in the above with $<$?}

By Eq. \ref{eq:mono}, the model will assign a larger probability to samples with larger FOM values.
With the fixed decoder $p_\theta(\hh{x} \vert \bm{z})$, we expand the marginal of $p_\theta$ in Eq. \ref{eq:mono} to obtain $f(\hh{x}) \monotonic{X}{+} \sum_{\bm{z}}[ p_\theta(\hh{x} \vert \bm{z})q_\phi(\bm{z})]$.
As we will show, to isotonically couple the latent sampler $q_\phi$ to $f(\hh{x})$, we pick a sampler $q_\phi$ that produces samples with probabilities isotonic to a surrogate function $h_\phi(\bm{z})$, i.e., $h_\phi(\bm{z}) \monotonic{Z}{+} q_\phi(\bm{z})$ over the latent space $\mathcal{Z}$.
Then, we train the latent sampler's surrogate function $h_\phi$ to be isotonic with $f$ with respect to a marginal over the latent space to achieve 
\begin{align}
      f(\hh{x}) &\monotonic{X}{+} \sum_{\bm{z}}[ p_\theta(\hh{x} \vert \bm{z})h_\phi(\bm{z})] \nonumber \\ &\monotonic{X}{+} \sum_{\bm{z}}[ p_\theta(\hh{x} \vert \bm{z})q_\phi(\bm{z})] . \label{eq:appmono}
\end{align}
Training $q_\phi$ and $h_\phi$ under these isotonic conditions is difficult because the decoder is often optimization-agnostic, meaning it isn't a priori trained to favor designs with large FOM, making it easy to break the isotone requirement in the worst case.
However, if the pretrained decoder is deterministic, as is often the case, then we can marginalize Eq. \ref{eq:appmono} to be more tractable.

In practice, the decoder $p_\theta(\hat{x} \mid z)$ of the bAE is deterministic: 
each latent vector $z$ maps to a single decoded design $\hat{x} = \mathcal{D}_\theta(z)$. 
This greatly simplifies the isotonicity condition in Eq.~\ref{eq:appmono} as
the probability of a design under the model depends only on the probability that 
the sampler $q_\phi$ assigns to the latent vector that produces it; thus, improving the probability of high-FOM designs is equivalent to 
increasing $q_\phi(z)$ for latent vectors whose decoded structures have large FOM. 
Consequently, the isotonic relationships in Eq.~\ref{eq:appmono} reduce to a simpler form:
\begin{equation}
\label{eq:decodermono}
    f(\mathcal{D}_\theta(z)) \monotonic{Z}{+} h_\phi(z) 
    \monotonic{Z}{+} q_\phi(z),
\end{equation}
where $h_\phi$ is the surrogate function used to guide the latent sampler.
This deterministic-decoder formulation allows PearSAN to directly enforce that 
latent vectors producing higher-performing TPV designs receive higher sampling 
probability, without retraining or modifying the decoder.

Our focus is now on realizing the isotonic conditions in Eq. \ref{eq:decodermono}. 
by 1) choosing a distribution $q_\phi$ that is isotonic with a trainable surrogate model $h_\phi$ and 2) enforcing that the surrogate model $h_\phi$ is isotonic with the FOM values, i.e., $ f(\hh{x}) \monotonic{X}{+} h_\phi(\bm{z})$, both of which are satisfied with PearSAN.

\begin{algorithm}[t]
\caption{PearSAN with Pretrained Decoder}
\begin{algorithmic}[1]
    \State \textbf{Require:} Initial dataset $\Z^{(0)} = \{\bm{z}^{(i)}\}$
    \State \textbf{Require:} Pretrained decoder $\mathcal{D}:\mathcal{Z} \rightarrow \mathcal{X}$
    \State \textbf{Require:} Figure-of-merit $f: \mathcal{X} \rightarrow \mathbb{R}$
    \State \textbf{Require:} Iterations $\tau_{\text{max}}$.
    \For{$\tau \in [0, ..., \tau_{\text{max}} - 1]$}
        \State Compute  FOM for new data $f(\mathcal{D}(\bm{z}^{(i)})) : \forall \bm{z}^{(i)} \in \Z^{(\tau)} $
        \State Retrain surrogate model $h_{\phi^{(\tau)}} \leftarrow \mathcal{L}_{PearSOL}(F^{(\tau)}, H^{(\tau)})$
        \State Sample surrogate model (e.g., VCA) $\hat{Z}^{(\tau)} \leftarrow \text{min}(h_{\phi^{(\tau)}}(T))$
        \State Update Dataset $\Z^{(\tau + 1)} \leftarrow \Z^{(\tau)} \cup \hat{Z}^{(\tau)}$ 
    \EndFor
    \State \textbf{Output:} Optimized design $\X^{(\tau_{\text{max}})} = \{\mathcal{D}_\theta(\bm{z}^{(i)}) : \bm{z}^{(i)} \in Z^{(\tau_\text{max})}\}$.
\end{algorithmic}
\label{alg:PearSAN}
\end{algorithm}

\subsection{Variational Neural Annealing}
\label{sec:VCA}
The most popular samplers $q_\phi$ which aim to produce samples isotonic with a surrogate function are Markov Chain Monte Carlo \cite{wilson_non-native_2024}, simulated annealing \cite{kirkpatrick_optimization_1983}, quantum samplers \cite{boixo_evidence_2014}, and more recently variational neural annealing  \cite{hibat-allah_variational_2021}. 
Inspired by statistical mechanics, these samplers are {\it antitonic} to an energy model  $h_\phi(\bm{z}) \monotonic{Z}{-} q_\phi(\bm{z}) $ where $\mathcal{Z} = \{0, 1\}^n$  and $h_\phi$ is given by a pseudo-boolean polynomial, 
\begin{align}
  h_\phi(\bm{z}) &= \sum_{s \subseteq {n}}\phi_s\prod_{i \in s}z_i \nonumber \\
       &= \sum_i \phi_iz_i + \sum_{i < j} \phi_{i,j}z_iz_j + \sum_{i < j < k} \phi_{i,j,k}z_iz_jz_k.... \label{eq:pubo}
\end{align}
typically representing the potential interactions between spin sites \cite{lucas_ising_2014}, superconducting rings \cite{boixo_evidence_2014}, etc. 

To construct $q_\phi(\bm{z})$, we implement variational classical annealing (VCA), a variant of variational neural annealing. 
VCA uses recurrent neural networks (RNNs) to sample $h_\phi(\bm{z})$ with better practical convergence compared to Markov Chain Monte Carlo and simulated annealing \cite{hibat-allah_variational_2021}, as we corroborate in Supplementary Information Section S5.
RNNs are widely used for generating sequential data, such as language modeling \cite{de_mulder_survey_2015}, anomaly detection \cite{basora_recent_2019}, and biometric authentication \cite{ackerson_applications_2021} through autoregression.
Each bit $z_i$ in the latent vector $\bm{z}$ is generated by a conditional probability statement $q_\phi(z_i \vert z_1,...,z_{i-1})$ following a Bernoulli distribution. VCA works through {\it minimizing} the sampler's variational free energy
\begin{equation}
    G_{\phi}(t) = \mathbb{E}_{\bm{z} \sim q_\phi(\bm{z})}[h_\phi(\bm{z})] - T(t) S(q_\phi),
\end{equation}
\bcl{Similar issue to above: we have $\lambda$, $\phi$, and $\phi$ as competing parameters sets--how do they related?}
where $T(t)$ is a temperature parameter that is decreased through the annealing process from a large $T(0) = T_0$ to $0$, and $S(q_\phi) = - \sum_{\bm{z}} q_\phi(\bm{z}) \log (q_\phi(\bm{z}))$ is the entropy.
We approximate the variational free energy at temperature $T$ as
\begin{equation}
    G_{\phi}(T) \approx \frac{1}{N_s} \sum_{i=1}^{N_s} h_\phi(\bm{z}^{(i)}) + T\log (q_\phi(\bm{z}^{(i)})), \label{eq:vfenergy}
\end{equation}
by taking $N_s$ discrete samples drawn from the RNN, i.e., $\bm{z}^{(i)} \sim q_\phi$.
The $T\log (q_\phi(\bm{z}^{(i)}))$ term has the effect of enforcing more randomness in the state evolution during the beginning of training with a high $T$, allowing the model to escape local minima and increasing the entropy throughout the annealing process. 
As $T$ decreases, analogous to simulated annealing, the model is less likely to exhibit large, random changes of state. To allow for the antitonicity of VCA, we replace the isotonic relations in Eq. \ref{eq:decodermono} with antitonic relations while still maintaining the overall objective, i.e.,
\begin{align}
      f(\mathcal{D}_\theta(\hh{z})) \monotonic{Z}{-} h_\phi(\bm{z}) \label{eq:fhiso} \\ h_\phi(\bm{z}) \monotonic{Z}{-}  q_\phi(\bm{z}) \label{eq:hqiso} \\
      \implies f(\mathcal{D}_\theta(\hh{z})) \monotonic{Z}{+} q_\phi(\bm{z}),
      \label{eq:appantiiso}
\end{align}
where Eq. \ref{eq:hqiso} is satisfied by using VCA.
To enforce Eq. \ref{eq:fhiso}, we introduce PearSOL.

\subsection{Pearson Surrogate Optimization Loss (PearSOL)}
\label{sec:pearsol}
Training an antitonic surrogate model typically involves modifying the {\it energy matching} (EM) loss, which uses a pairwise L-norm loss function \cite{wilson_machine_2021}, e.g.,
\begin{equation}
    \mathcal{L}_{\text{EM}}(F, H) = \sum_{i}\left \lVert F_i - (-H_i) \right \rVert_2^2,\label{eq:energymatch}
\end{equation}
where $F = \{f(\mathcal{D}_\theta(\bm{z}^{(i)}))\}_{i=1}^n$ represents the set of decoded FOMs and $H = \{h_\phi(\bm{z}^{(i)})\}_{i=1}^n$ represents the latent vector energies. EM minimizes the difference between $-H^{(i)}$ and $F^{(i)}$, thus achieving antitonicity, but L-norm loss can be overly sensitive in less-explored regions and unnecessarily penalizes small differences. 
To overcome these issues and enforce antitonicity, we propose using Pearson correlation \cite{benesty_pearson_2009, lu_pearnet_2022}:
\begin{equation}
    \mathcal{L}_\text{Pearson}(F, H) = \frac{\sum_i (F_{i} - \overline{F})(H_{i} - \overline{H})}{\hat{S}_F \hat{S}_H}
\end{equation}
where $\overline{H}$ and $\hat{S}_H$ are the mean and standard deviation of $H$, and $\overline{F}$ and $\hat{S}_F$ are defined similarly for $F$.
By the Cauchy-Schwarz inequality, $\mathcal{L}_\text{Pearson}(F, H) \in [-1, 1]$, with $+1$ or $-1$ indicating perfect isotonic or antitonic correlations, respectively (See Supplementary Information Section S2).
Our goal is to achieve a Pearson correlation of $-1$, achieving antitonicity $f(\mathcal{D}(\bm{z})) \monotonic{Z}{-} h_\phi(\bm{z})$. 
In practice, gradient convergence requires hyperparameter tuning and an additional inverse logistic curve (see Supplementary Information, Section S3).

To improve the antitonicity and the lower-energy distribution of points, we impose an additional regularization loss in the form of an average energy loss $\mathcal{L}_\text{Avg}(H) = \overline{H}$ which simply minimizes the average energy.
However, to ensure that the surrogate model's parameters remain near $1$ and maintain the norm of $h_\phi$, we add a regularization term $\mathcal{L}_\text{Norm}$, lest $\mathcal{L}_\text{Avg}$ is satisfied through exploding negative energy function parameters.
Our full PearSOL function reads,
\begin{equation}
    \mathcal{L}_\text{PearSOL} = \lambda_a \mathcal{L}_\text{Pearson} + \lambda_b \mathcal{L}_\text{Avg} + \lambda_c \mathcal{L}_\text{Norm},\label{eq:loss_corr}
\end{equation}
with hyperparameters $\lambda_a, \lambda_b, \lambda_c$ and each loss implicitly using $H$ and $F$ where appropriate.

\begin{figure*}[t!]
    \centering
    \includegraphics[width=\textwidth]{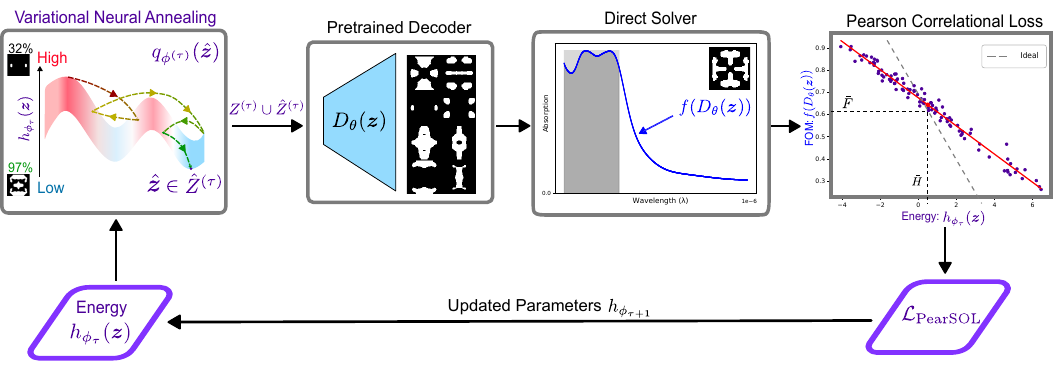} 
    \caption{{\bf PearSAN } starts with an initial dataset $Z^{(0)}$. The generated polynomial $h_{\phi^{(\tau)}}(\bm{z})$ is then used in variational neural annealing to train the sampler $q_{\phi^{(\tau)}}(\bm{z})$ by minimizing its free energy $F_{\phi^{(\tau)}}(t)$. The resultant latent vectors $\hat{\bm{z}}$ are then stored in a database. The sampler is then evaluated through the pretrained decoder, which generates real samples $D_{\theta}(\hat{\bm{z}})$. The samples' efficiency is computed using a direct solver, as indicated by the grey area between the ideal and resultant spectra. The efficiency $f(D_{\theta}(\hat{\bm{z}}))$, along with its corresponding sample, are also stored in the database. The Pearson correlation loss is informed through the database, which constitutes an antitonic correlation between $h_{\phi^{(\tau)}}(\bm{z})$ and $f(D_{\theta}(\hat{\bm{z}}))$. The resulting loss value from PearSOL is used to update the surrogate model parameters in the next iteration $\phi^{(\tau + 1)}$, and the process is repeated.}
    \label{fig:main}
\end{figure*}

\subsection{PearSAN Overview}
\label{sec:PearSAN}
The overall workflow of PearSAN is shown in Figure \ref{fig:main}. PearSAN is especially effective when the initial dataset $X^{(0)}$ is updated with new designs generated by the decoder, allowing the surrogate model to retrain on the FOM from the new designs. 
We denote the current iteration with $\tau$ and the total iterations with $\tau_\text{max}$.
As outlined in Alg. \ref{alg:PearSAN}, we begin by sampling an initial set of latent vectors $Z^{(0)} = \{\bm{z}^{(i)}\}$ either via an encoder $\bm{z}^{(i)} \sim \mathcal{E}(\bm{x}^{(i)}) : \bm{x}^{(i)} \in X^{(0)}$ or some prior $\bm{z}^{(i)} \sim q(\bm{z})$.
Each iteration begins by training a polynomial surrogate model $h_{\phi^{(\tau)}}$ (Eq. \ref{eq:pubo}) to optimize the PearSOL over $Z^{(\tau)}$, $\mathcal{L}_\text{PearSOL}(F^{(\tau)}, H^{(\tau)})$ where 
\begin{align}
  H^{(\tau)} &= \{h_{\phi^{(\tau)}}(\bm{z}^{(i)}) : \bm{z}^{(i)} \in Z^{(\tau)}\}  \\
  F^{(\tau)} &= \{f(\mathcal{D}_\theta(\bm{z}^{(i)})) : \bm{z}^{(i)} \in Z^{(\tau)}\},
\end{align}
and $\mathcal{D}_\theta$ is a pretrained, deterministic decoder $\mathcal{D}_\theta(\bm{z})$ with a discrete latent space.
Then, we train a RNN $q_{\phi^{(\tau)}}$ to antitonically sample the surrogate model $h_{\phi_{\tau}}$ using VCA by minimizing Eq. \ref{eq:vfenergy}. 
As evidenced by the success of dropout, score matching, and diffusion models, adding noise during training can promote exploration and improve accuracy in low-probability regions of the latent space. 
For VCA, the high-temperature $T = T_0$ noisy sampling at the beginning of training serves both to aid the annealing process and to increase the breadth of sampled latent vectors.
Therefore, we accumulate latent vectors $\hat{Z}^{(\tau)}$ throughout the noisy VCA steps to allow the surrogate model to learn both breadth and depth of the latent space's relation to the FOM.
Practically, we set an epoch cutoff $N_{\text{thresh}}$, before collecting latent vectors for training to exclude vectors generated at excessively high temperatures which are less responsive to $h_{\phi^{(\tau)}}$.
We accumulate the new latent vectors $\hat{Z}^{(\tau)} = \{\hh{z}^{(i)}\} : \hh{z} \sim q_{\phi^{(\tau)}}(\hh{z})$ and construct a new dataset $Z^{(\tau + 1)} = Z^{(\tau)} \cup \hat{Z}^{(\tau)}$. We repeat this procedure a total of $\tau_\text{max}$ times.

\section{Result and Discussion}
\label{sec:results}
To empirically test PearSAN, we compare the reported designs and runtimes against existing optimization methods for the TPV design problem. 
Additionally, we perform an ablation study to understand the importance of PearSOL for PearSAN.
We demonstrate that for training a surrogate model, PearSOL is preferable to EM.
For all loss-function comparisons, we use the same hyperparameters, TPV unit-cell dataset, architectures for the bAE and RNN, and the optimizer's temperature schedule.

To quantify the degree of information loss introduced by using the bAE as the encoder–decoder model, we report reconstruction metrics (MSE and perceptual loss) and sample topology comparisons in Supplementary Information, Section~S4.1 (Figure~S7). 
We also provide distribution-level fidelity metrics (KID, FID, and Inception Score) in Section S4.2. 
The inclusion of the perceptual loss term substantially improves structural reconstruction quality, producing sharp and nearly identical layouts relative to the original designs.

In Supplementary Information Section S1.2, we attempt to upgrade EM with positive affine parameters over the surrogate model to help EM  yield similar results to PearSOL.
However, we find that the affine parameters degrade training performance, likely due to difficulties with fine-tuning.
Superior quality in both decoder scores and optimization quality reinforces our choice of PearSOL for PearSAN.

\begin{figure*}[h]
\centering
\includegraphics[width=0.8\textwidth]{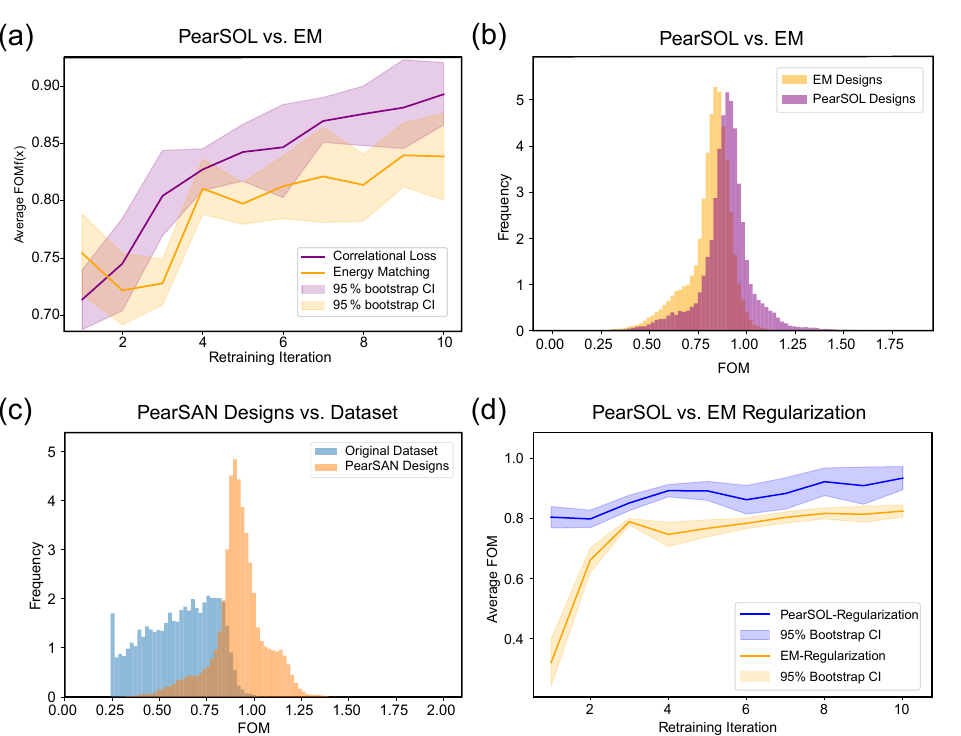}
    \caption{\textbf{Retraining performance for PearSOL and EM:} (a) Average sampled FOM over 10 retraining iterations, (b) FOM histogram of all decoded vectors from the last iteration, (c) Comparison of FOM between vectors generated by PearSAN and the original dataset (unnormalized VGGNet FOM), (d) Regularization performance of PearSAN versus EM, comparing sampling performance of one binary autoencoder trained with PearSOL vs. one with EM. Bootstrapped confidence intervals are calculated with 10,000 samples \cite{agarwal2021deep}. We use PearSOL for sampling, but each decoder is separately trained with either an additional PearSOL or EM term.}
    \label{fig:retraining}
\end{figure*}

\begin{figure*}[h]
    \centering
    \includegraphics[width=0.8\textwidth]{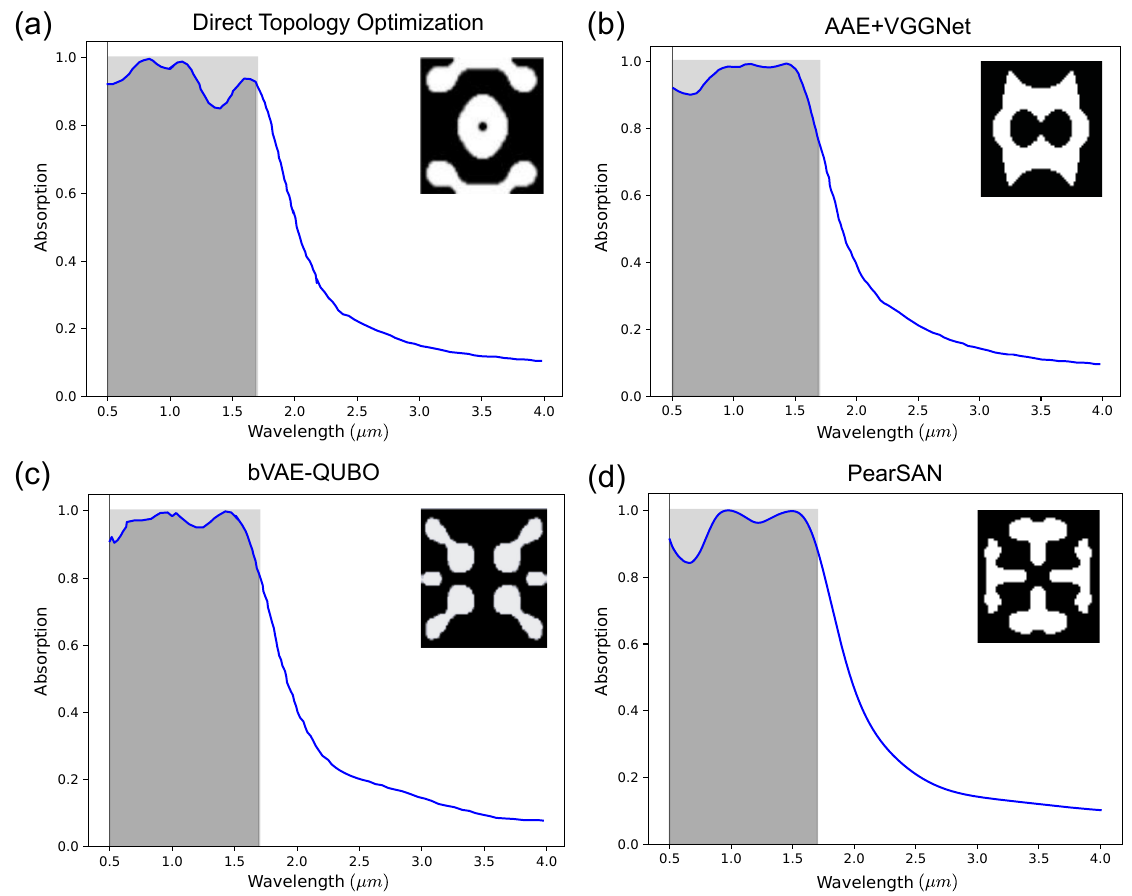}
    
    \caption{\textbf{Spectra Comparison of different optimization methods} PearSAN consistently stays within high values of absorption. NOTE: Sub-figures (a), (b), (c) adapted with permission from \cite{kudyshev_machine-learning-assisted_2020} and \cite{wilson_machine_2021} respectively.}
    \label{fig:spectra_comp}
\end{figure*}

\begin{table}[H]
    \centering
    \begin{tabular}{l|c|c|c|c}
    \toprule
    Method & Efficiency (\%)& Hours per 100 designs& Designs per minute& Parameter count\\
    \midrule
    Direct Topology Optimization \cite{kudyshev_machine-learning-assisted_2020} & 92.00 & 164 & 0.0102 & -\\
    AAE + VGGNet \cite{kudyshev_machine-learning-assisted_2020} & 95.50 & 0.0333 & 50.1 & 4.3 million \\
    AAE + TO*\cite{kudyshev_machine-learning-assisted_2020} & 97.90 & 54 & 0.031 & 4.3 million \\
    AAE + DE \cite{kudyshev_machine_2021} & 95.90 & 23.33 & 0.0714 & 8.4 million \\
    AAE + rDE \cite{kudyshev_machine_2021} & 96.40 & 23.33 & 0.0714 & 8.4 million\\
    bVAE-QUBO \cite{wilson_machine_2021} & 96.70 & 0.30 & 5.556 & 7.9 million \\
    \textbf{PearSAN (PearSOL)} & \textbf{97.02} & \textbf{0.0033} & \textbf{501.10} & 7.7 million \\
    \bottomrule
    \end{tabular}
    \caption{\textbf{Comparison of PearSAN (PearSOL) with previous optimization methods in terms of design efficiency and sampling time:}
    The proposed method shows a significant inference time speed improvement while achieving the highest efficiency outside of AAE+TO. However, AAE+TO and Direct Topology Optimization utilized FDTD simulations to perform adjoint optimization for fine-tuning, rather than the VGG model employed in most other methods.}
    \label{tab:speed_comparison}
\end{table}

\subsection{Retraining Procedure}
We consider $\tau_{\text{max}} = 10$ iterations of PearSAN and a sampling epoch threshold at $N_{thresh} = 20$ for VCA, as we observe diminishing returns for larger values.
For each retraining iteration, we compute the average FOM across all accumulated vectors from $10$ experiments, excluding the first iteration, which is undertrained due to random initialization.
Figure \ref{fig:retraining} (a) shows that PearSOL consistently outperforms EM
for all retraining iterations, achieving an average of 92.31\% VGGNet-predicted efficiency on the final retraining iteration $\tau_{\text{max}} - 1$, versus energy matching with an average of 80.74\%. In the experiment of Figure \ref{fig:retraining} (b), we conduct a Welch two-sample t-test comparing the mean FOM values from the final iteration $\tau_{\text{max}}$ of PearSOL (85.81\%) versus EM (79.08\%), finding that PearSOL significantly outperformed EM ($t(607375) = 203.51, p < 2.2 \times 10^{-16}$). In Figure \ref{fig:retraining} (c), we show how our technique significantly improves upon the quality of our dataset (mean 92.32\% versus 60.54\%). Furthermore, we consider both losses on two bAEs, one regularized with PearSOL and one with EM loss with affine parameters, discussed further in Section S1.2 of the Supplementary Information.


\subsection{Comparison with Previous Methods}
\label{sec:results_previous}
We compare PearSAN against previous methods such as direct topology optimization \cite{kudyshev_machine-learning-assisted_2020}, AAE+TO, AAE+VGGNet \cite{kudyshev_machine-learning-assisted_2020}, AAE+DE, AAE+rDE \cite{kudyshev_machine_2021} and bVAE-QUBO \cite{wilson_machine_2021} based on efficiency, time taken (in hours) to produce 100 designs, designs generated per minute, and the number of parameters per model. This is also shown in Table \ref{tab:speed_comparison}.
To find the optimal design, we took the 100 best designs from the best run of PearSAN with PearSOL and simulated them using the same finite-difference time-domain methods as the other techniques. While previous methods evaluated 100 designs over several hours, PearSAN was able to evaluate 100 designs in just $0.0033$ hours, achieving high computational efficiency and rapid sampling speeds prioritized in recent surrogate benchmarks \cite{augenstein_neural_2023}. PearSAN's optimal design and its corresponding spectra are shown in Figure \ref{fig:spectra_comp} (d). PearSAN consistently stays at higher levels of absorption within the working band, only starting to dip outside the band at an absorption rate of 0.85, compared to that of AAE+VGGNet's 0.75 and bVAE-QUBO's 0.8 (Figure \ref{fig:spectra_comp} (b) and (c)). Further, PearSAN is also more consistently at maximum absorption compared to direct topology optimization (Figure \ref{fig:spectra_comp} (a)).
PearSAN achieved the highest efficiency of $97.02\%$, outperforming all prior approaches which were limited by the VGGNet for FOM prediction \footnote{We include a precision of $10^{-2}$ for comparison purposes even though fabrication will introduce errors of $\pm 5\%$.}. Notably, PearSAN only had access to the VGGNet, while AAE+TO had direct access to FDTD calculations during training. Thus, a more appropriate comparison is with AAE+VGGNet, which is architecturally equivalent to AAE+TO but constrained by the same FOM approximation. As such, PearSAN's superior performance shows that, given a better FOM approximation network or direct access to FDTD simulations, PearSAN could surpass AAE+TO.

\section{Conclusion and Outlook}

In this work, we introduce PearSAN, a machine-learning–assisted optimization framework for inverse design problems in high-dimensional parameter spaces. Applied to the metasurface unit-cell optimization task for TPVs, PearSAN combines PearSOL with VCA sampling to exploit discretized latent spaces, accelerating search while improving design quality.

By comparing PearSOL against conventional energy-matching surrogate losses, we showed that EM produces suboptimal latent geometries even when augmented with affine corrections. Across 10 retraining iterations, PearSOL consistently achieved higher VGGNet-predicted average FOMs (92.31\% vs. 80.74\%) and superior generative performance as measured by KID, FID, and Inception Score. Furthermore, we compare our method with prior ML-based approaches and direct topology optimization, demonstrating that it achieves higher design efficiency in significantly less time. We additionally introduced a practical application of VCA to sample topologies.
Overall, PearSAN reached a state-of-the-art maximum design efficiency of $\sim 97\%$ with a mean of $\sim 71\%$ while offering at least an order-of-magnitude speedup relative to competing baselines.

These results highlight PearSAN as a powerful and flexible tool for accelerating inverse design pipelines. Its combination of rapid sampling and robust surrogate modeling, without requiring retraining of the outer autoencoder, makes it well-suited to a broad class of complex, physics-constrained optimization tasks.

\subsection{Future Directions}

While we demonstrate PearSAN with a quadratic Boolean energy function, different physics-inspired energy models (e.g., Blume-Capel, Potts, or higher-order polynomials) may offer richer representations of the design space. 
Additionally, PearSAN can be adapted to continuous latent spaces, utilizing gradient-based search.
Beyond autoregressive models for VCA, exploring different latent-space optimizers or advanced neural architectures such as State Space Machines \cite{gu2024mambalineartimesequencemodeling} could further accelerate convergence.
Beyond the TPV problem, the application of PearSAN to various nanophotonic device design tasks, such as quantum information and sensing, will help elucidate its generality and performance benefits.

Although our experiments already exhibit speed differences spanning several orders of magnitude, a fully controlled comparison on identical hardware would enable future efforts to disentangle algorithmic gains from implementation-level effects.

\medskip
\textbf{Supplementary Information} \par 
Supplementary Information is available from the Wiley Online Library or from the author.

\medskip
\textbf{Acknowledgements} \par 
This work is supported by the Purdue’s Elmore ECE Emerging Frontiers Center ‘The Crossroads of Quantum and AI’, and National Science Foundation (NSF) award DMR-2323910.

\medskip

%

\bibliographystyle{MSP}
\bibliography{main}

\end{document}